\documentclass[journal]{IEEEtran}

\usepackage{orcidlink}
\usepackage{amsmath,amssymb,amsfonts}
\usepackage{algorithmic}
\usepackage{graphicx}
\usepackage{textcomp}
\usepackage{float}
\usepackage{caption}
\usepackage{multirow}
\usepackage{mathtools}
\usepackage{bm}
\usepackage{adjustbox}
\usepackage{subfigure}
\usepackage[table]{xcolor}
\usepackage{soul}
\usepackage{gensymb}
\usepackage{authblk}
\usepackage[utf8]{inputenc}
\usepackage{booktabs}
\usepackage{nameref}
\usepackage{placeins}
\usepackage[ruled,vlined]{algorithm2e}
\SetKw{Return}{return}
\usepackage{hyperref}
\usepackage{siunitx} 
\usepackage[T1]{fontenc}

\ifCLASSINFOpdf
\else
\fi

\begin{document}

\title{Enhanced INS/GNSS State Estimation using GNSS-Based Acceleration Measurements}

\author{Gal~Versano~\orcidlink{0009-0007-7766-2511},
        Itzik~Klein~\orcidlink{0000-0001-7846-0654}
        
  \thanks{G. Versano and I. Klein are with the Autonomous Navigation and Sensor Fusion Lab, Hatter Department of Marine Technologies, Charney School of Marine Sciences, University of Haifa, Israel.}}
\maketitle

\begin{abstract}
Accurate and reliable navigation is essential for autonomous ground vehicle operations. Standard INS/GNSS fusion relies on GNSS position updates, which provide limited observability of orientation and inertial sensor error states, particularly during low-dynamic motion. In this work, we propose utilizing past GNSS measurements alongside a motion model to extract meaningful vehicle acceleration information. This acceleration measurement is then integrated into the INS/GNSS filter to improve its robustness and accuracy. The proposed approach is evaluated on two real-world unmanned ground vehicle datasets collected from different mobile platforms and inertial sensor grades. Results demonstrate consistent positioning accuracy improvements relative to the standard position-aided filter, with mean position root mean square error improvements of 11.40\% and 20.74\% on the two datasets, respectively.
\end{abstract}

\begin{IEEEkeywords}
Inertial Navigation System, Global Navigation Satellite Systems, Ground Vehicle Navigation, Extended Kalman Filter
\end{IEEEkeywords}

\section{Introduction}
\noindent The rapid development of autonomous ground vehicles is fueled by the promise of transformed operational efficiency and industrial innovation \cite{basiuk2023mobile, raj2022comprehensive,antonyshyn2023multiple}. These systems are increasingly integrated across specialized sectors: precision agriculture utilizes them for automated harvesting, while logistics centers rely on them for streamlined warehouse management. Furthermore, they play a critical role in high-stakes environments, including search-and-rescue missions and autonomous underground exploration \cite{yepez2023mobile,lin2023intelligence,neaz2023design}.\\
\noindent Reliable and precise navigation is the cornerstone of effective unmanned ground vehicle (UGV) operations. To ensure high-frequency state estimation and rapid system responsiveness, most autonomous platforms fuse data from inertial sensors with external updates. These multi-modal updates typically leverage technologies such as inertial navigation systems (INS), global navigation satellite systems (GNSS), vision-based localization, and light detection and ranging (LiDAR) \cite{yue2024lidar,patoliya2022robust,li2022tightly}. \\
\noindent AS the INS/GNSS fusion process is nonlinear, commonly the extended Kalman filter (EKF) is employed~\cite{bar2001estimation, groves2015principles}.
The EKF addresses nonlinearity by approximating the state transition and measurement models through first-order Taylor series expansions. However, this linearization around the current estimate can introduce notable errors in both the predicted mean and covariance, potentially causing instability or divergence when the system exhibits strong nonlinear behavior~\cite{brigadnov2023error}. Nevertheless, the EKF remains popular in robotics and navigation because it offers a good balance between computational efficiency and reliable tracking of complex system dynamics~\cite{borko2018gnss,jiang2022effective,wang2019constrained}. Other variants of the EKF, applicable for sensor fusion, are the invariant Kalman filter (IKF) \cite{barrau2018invariant, diker2026neural, xia2024invariant}, or unscented Kalman filter (UKF)~\cite{al2021novel,meng2016covariance,levy2026adaptive}. 
\noindent Information aiding aims to exploit motion constraints and environmental interactions to serve as external updates, providing timely corrections to the navigation filter~\cite{engelsman2023information}. One type of such aiding leverages historical sensor data to provide auxiliary updates to the EKF. For instance, Klein~\cite{klein2020squeezing} utilized past GNSS measurements to derive orientation updates for the navigation filter. Later, a similar approach was used to generate velocity updates during complete Doppler velocity log (DVL) outages by relying on previously recorded DVL data~\cite{klein2020continuous}. An extension of this work suggested using past DVL measurements to estimate a DVL acceleration-based update~\cite{11244966}.\\
\noindent In standard INS/GNSS fusion, position updates provide limited information about the orientation and inertial error states. These states become observable mainly during dynamic maneuvers, leaving them poorly estimated during mild or steady motion \cite{farrell2008aided}.  To bridge this gap, this work proposes utilizing past GNSS position measurements and a motion model to extract meaningful vehicle acceleration information. This acceleration measurement is then introduced into the INS/GNSS filter to improve its robustness and accuracy. Because the acceleration measurement model directly incorporates the transformation matrix and accelerometer bias errors, it offers a more direct means of estimating these states. By extracting an acceleration estimate from existing GNSS position data, requiring no additional hardware, and feeding it into the navigation filter as an auxiliary update, orientation and bias estimation can be improved regardless of the vehicle's motion profile.\\
\noindent
The main contributions of this paper are as follows:
\begin{enumerate}
	\item A least-squares approach to derive a smooth acceleration estimate 
	directly from raw GNSS position measurements.
	\item An acceleration measurement model that integrates this GNSS-derived 
	acceleration as an auxiliary EKF update alongside standard position updates.
\end{enumerate}

\noindent Our proposed approach was evaluated  on two real-world unmanned ground vehicle datasets, collected from different mobile platforms and inertial sensor grades. Results demonstrate consistent positioning accuracy improvements relative to the position-aided filter, with a mean position root mean square error improvement of 11.40\% and 20.74\% on the two datasets, respectively.\\
\noindent The rest of the paper is structured as follows: Section \ref{prob_for} provides a detailed explanation of the problem formulation. Section \ref{propose_app} presents our proposed approached. Section \ref{res} presents our analysis and results and \ref{conc} gives the conclusions of this work.\\
\section{Problem Formulation}\label{prob_for}
\subsection{INS Kinematic Equations}
\noindent The navigation equations of motion, expressed in the navigation frame ($n$), describe the evolution of position, velocity, and orientation.  The position vector is expressed in geodetic coordinates $p^n = [L, \lambda, h]^T$, with its rate of change given by:
\begin{equation}
\begin{bmatrix} \dot{L} \\ \dot{\lambda} \\ \dot{h} \end{bmatrix} = \begin{bmatrix} \frac{v_N}{R_M + h} \\ \frac{v_E}{(R_N + h)\cos L} \\ -v_D \end{bmatrix}
\end{equation}
where $R_M$ and $R_N$ represent the meridian radius and normal radius of curvature, respectively.\\
\noindent The rate of change of the velocity vector is expressed as:
\begin{equation}\label{eq_ins1}
\boldsymbol{\dot{v}}^{n}=\mathbf{R}^{n}_{b}\boldsymbol{f}^{b}_{ib}+\boldsymbol{g}^{n}-\left(\mathbf{\Omega}^{n}_{en}+2\mathbf{\Omega}^{n}_{ie}\right) \boldsymbol{v}^{n}
\end{equation}
where $f^b_{ib}$ is the specific force measured by accelerometers, $R_b^n$ is the transformation matrix from the body frame to the navigation frame, $g^n$ is the gravity vector, $\boldsymbol{\Omega}_{en}^n$ and $\boldsymbol{\Omega}_{ie}^n$ represent the transport rate and earth rotation rate skew matrix, respectively.
The transformation matrix from the body frame to the navigation frame evolves as:
\begin{equation}\label{eq_ins2}
\dot{\mathbf{R}}^{n}_{b}=\mathbf{R}^{n}_{b}\mathbf{\Omega}^{b}_{ib}-\left(\mathbf{\Omega}^{n}_{ie}+\mathbf{\Omega}^n_{en}\right)\mathbf{R}^{n}_{b}
\end{equation}
where $\Omega_{ib}^b$ is the skew-symmetric form of the angular rate measured by the gyroscope.
\subsection{Navigation Filter}
\noindent Due to the nonlinear nature of the INS equations, an error-state EKF is implemented. The 15-state error vector $\delta x$ is defined as:
\begin{equation} \label{eq:errorState}
\boldsymbol{\delta x} = \left[ \begin{array}{ccccc}
\boldsymbol{\delta p^{n}} & \boldsymbol{\delta v^{n}} & \boldsymbol{\phi^{n}} & \boldsymbol{b_{a}} & \boldsymbol{b_{g}} \end{array} \right]^{T} 
\end{equation}
\noindent where $\boldsymbol{\delta p^{n}}$ is the position error-states, $\boldsymbol{\delta v^{n}}$ is the velocity error-states, $\boldsymbol{\phi^{n}}$ is the misalignment error-states, and $\boldsymbol{b_{a}}$ and $\boldsymbol{b_{g}}$ are the accelerometer and gyroscope bias residual errors, respectively. \\
\noindent The linearized error state dynamic model is expressed as:
\begin{equation}
\boldsymbol{\delta \dot {x}} =\mathbf{F} \boldsymbol{\delta x} +\mathbf{G}\boldsymbol{w} 
\end{equation}
\noindent where $\boldsymbol{F}$ is the system matrix
\begin{equation}
\mathbf{F} = \begin{bmatrix} 
\mathbf{F}_{pp} & \mathbf{F}_{pv} & \mathbf{0}_{3 \times 3} & \mathbf{0}_{3 \times 3} & \mathbf{0}_{3 \times 3} \\
\mathbf{F}_{vp} & \mathbf{F}_{vv} & \mathbf{F}_{v\psi} & \mathbf{R}_b^n & \mathbf{0}_{3 \times 3} \\
\mathbf{F}_{\psi p} & \mathbf{F}_{\psi v} & \mathbf{F}_{\psi \psi} & \mathbf{0}_{3 \times 3} & \mathbf{R}_b^n \\
\mathbf{0}_{3 \times 3} & \mathbf{0}_{3 \times 3} & \mathbf{0}_{3 \times 3} & \mathbf{0}_{3 \times 3} & \mathbf{0}_{3 \times 3} \\
\mathbf{0}_{3 \times 3} & \mathbf{0}_{3 \times 3} & \mathbf{0}_{3 \times 3} & \mathbf{0}_{3 \times 3} & \mathbf{0}_{3 \times 3}
\end{bmatrix},
\end{equation}
$\boldsymbol{G}$ is the shaping matrix
\begin{equation}
\boldsymbol{\mathrm{G}} =~\left[ \begin{array}{cccc}
\boldsymbol{0_{3\times3}} & \boldsymbol{0_{3\times3}} & \boldsymbol{0_{3\times3}} & \boldsymbol{0_{3\times3}} \\ 
\mathbf{R}^{n}_{b} & \boldsymbol{0_{3\times3}} & \boldsymbol{0_{3\times3}} & \boldsymbol{0_{3\times3}} \\ 
\boldsymbol{0_{3\times3}} & \mathbf{R}^{n}_{b} & \boldsymbol{0_{3\times3}} & \boldsymbol{0_{3\times3}} \\ 
\boldsymbol{0_{3\times3}} & \boldsymbol{0_{3\times3}} & \mathbf{{I}_{3}} & \boldsymbol{0_{3\times3}} \\ 
\boldsymbol{0_{3\times3}} & \boldsymbol{0_{3\times3}} & \boldsymbol{0_{3\times3}} & \mathbf{{I}_{3}} \end{array}
\right],
\end{equation}
$\boldsymbol{w}$ is the noise vector
\begin{equation}
\boldsymbol{w} = \left[ \begin{array}{cccc}
\boldsymbol{w_{a}} & \boldsymbol{w_{g}} & \boldsymbol{w_{a_{b}}} & \boldsymbol{w_{g_{b}}} \end{array} \right]^{T}
\end{equation}
$\boldsymbol{w_{a}}$ and $\boldsymbol{w_{g}}$ are the accelerometer and gyroscope measurements white noise, respectively, and $\boldsymbol{w_{a_{b}}}$ and $\boldsymbol{w_{g_{b}}}$ are the accelerometer and gyroscope biases white noise, respectively.
\noindent The Kalman filter predicts the state and its uncertainty and then updates them using external measurements, where the Kalman gain sets the relative weighting between prediction and measurement. The error state covariance is propagated as:
\begin{equation}\label{eq_p}
\mathbf{\hat{P}}^{-}_{k} = \mathbf{\Phi}_{k-1} \mathbf{\hat{P}}^{+}_{k-1} \mathbf{\Phi}_{k-1}^{T} + \mathbf{Q}_{k-1}
\end{equation}
 where $\mathbf{\hat{P}}^{-}_{k}$ and $\mathbf{\hat{P}}^{+}_{k}$ denote the predicted and updated covariance estimates, respectively, and $\mathbf{Q}_{k}$ is the process noise covariance matrix. \\
\noindent Since the INS dynamics are nonlinear, the state transition matrix $\mathbf{\Phi}$ is obtained from the linearized error-state model, commonly approximated using a first-order discretization:
\begin{equation}
\mathbf{\Phi} \approx \mathbf{I} + \mathbf{F} dt
\end{equation}
where $dt$ is the sampling interval time. \\
\noindent The Kalman gain is computed as:
\begin{equation}
\mathbf{K}_{k} = \mathbf{\hat{P}}^{-}_{k} \mathbf{H}_{k}^{T} \left( \mathbf{H}_{k} \mathbf{\hat{P}}^{-}_{k} \mathbf{H}_{k}^{T} + \mathbf{R}_{k} \right)^{-1}
\end{equation}
where $\mathbf{H}_{k}$ is the measurement matrix and $\mathbf{R}_{k}$ is its associated measurement noise covariance.\\
\noindent The updated covariance matrix is given by
\begin{equation}
\mathbf{\hat{P}}^{+}_{k} = \left( \mathbf{I} - \mathbf{K}_{k}\mathbf{H}_{k} \right)\mathbf{\hat{P}}^{-}_{k}
\end{equation}
\noindent Finally, the corrected error-state estimate in a closed-loop formulation is expressed as \cite{farrell2008aided}:
\begin{equation}
\boldsymbol{\delta \hat{x}}^{+}_{k} = \mathbf{K}_{k} \boldsymbol{\delta z}_{k}
\end{equation}
where $\boldsymbol{\delta z}_{k}$ is the measurement residual vector.
\begin{figure*}
    \centering
    \includegraphics[width=0.8\linewidth]{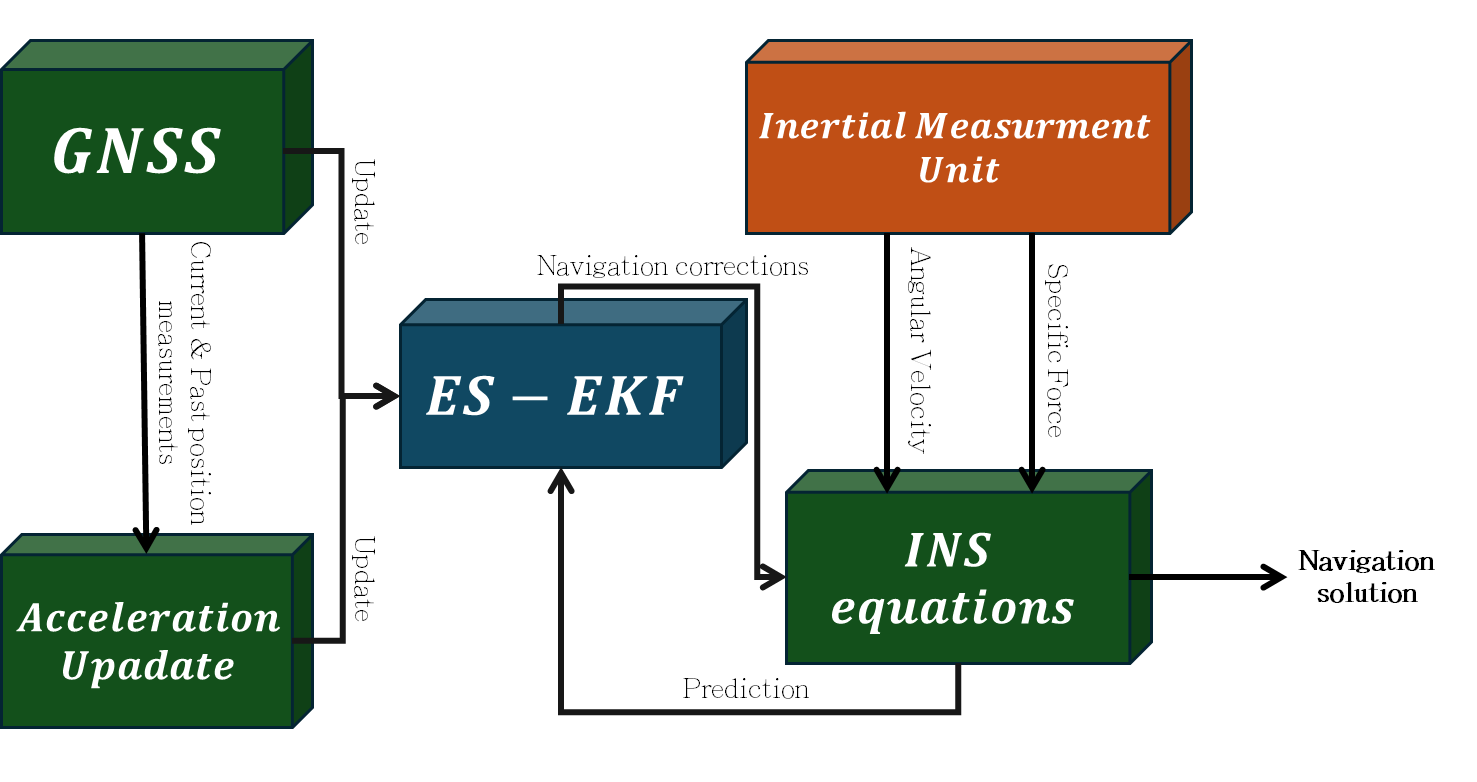}
    \caption{Flowchart of the proposed approach, integrating GNSS-based acceleration updates into the INS/GNSS navigation filter.}
    \label{scheme_acc_ekf_upd}
\end{figure*}
\section{Proposed Approach} \label{propose_app}
\noindent  
Instead of using only GNSS position updates, we  derive an acceleration update from past GNSS position measurements. By combining the acceleration measurement with the position measurement in the navigation filter, we aim to improve the INS/GNSS filter performance. A flow chart of the proposed approach is given in Figure \ref{scheme_acc_ekf_upd}. We note, that although the GNSS position and GNSS acceleration measurements are correlated, as the latter is constructed from past position history, this correlation is neglected in the current study.
\subsection{Accelerometer Information from past GNSS measurements}
\noindent To extract acceleration information from past GNSS position measurements, we employ a least squares (LS)  fitting approach as follows. Given $j$ past GNSS position measurements at time instances $t_j$, the vehicle motion can be approximated using a second-order Taylor expansion. To preserve causality, the model is referenced to the beginning of the current window, $t_0$:
\begin{equation}
\mathbf{p}(t_j) = \mathbf{p}_0 + \mathbf{v}_0 \Delta t_j + \frac{1}{2}\mathbf{a} \Delta t_j^2,
\qquad \Delta t_j = t_j - t_0
\label{eq:poly}
\end{equation}
\noindent Stacking all $m$ measurements yields the linear system is:
\begin{equation}    
\mathbf{A}\boldsymbol{\theta} = {\mathbf{p}}
\end{equation}
where the design matrix $\mathbf{A}$ and parameter vector $\boldsymbol{\theta}$ are defined as:
\begin{equation}
\mathbf{A} =
\begin{bmatrix}
1 & \Delta t_0 & \frac{1}{2}\Delta t_0^2 \\
1 & \Delta t_1 & \frac{1}{2}\Delta t_1^2 \\
\vdots & \vdots & \vdots \\
1 & \Delta t_{m-1} & \frac{1}{2}\Delta t_{m-1}^2
\end{bmatrix},
\quad
\boldsymbol{\theta} =
\begin{bmatrix}
\mathbf{p}_n \\
\mathbf{v}_n \\
{\mathbf{a}}^{n}
\end{bmatrix}
\label{eq:A}
\end{equation}
The LS minimizes the squared residual between the measured positions and the fitted solution by:
\begin{equation}
\begin{bmatrix}
\mathbf{p}_0 \\
\mathbf{v}_0 \\
\tilde{\mathbf{a}}^{n}
\end{bmatrix}
=
(\mathbf{A}^T \mathbf{A})^{-1} \mathbf{A}^T
\begin{bmatrix}
\tilde{\mathbf{p}}_0 \\
\tilde{\mathbf{p}}_1 \\
\vdots \\
\tilde{\mathbf{p}}_{m-1}
\end{bmatrix}
\label{eq:lstsq}
\end{equation}
 The estimated acceleration vector $\tilde{\mathbf{a}}^{n}$ can be directly extracted using  $\mathbf{B} \in \mathbb{R}^{1 \times m}$, yielding:
\begin{equation}
\tilde{\mathbf{a}}^{n} = \mathbf{B}\,{\mathbf{p}},
\qquad
\mathbf{B} = \text{row}_3 \left[(\mathbf{A}^T \mathbf{A})^{-1} \mathbf{A}^T \right]
\label{eq:acc}
\end{equation}
In our current implementation, we use a window size of three seconds corresponding to three past position measurements. Acceleration estimates are produced only after the buffer is fully populated, ensuring that each estimate is based on a consistent history of measurements.
\subsection{Accelerometer-Aided Measurement Model}
\noindent In this section, we derive the acceleration measurement model using GNSS-derived acceleration. The measurement residual is defined as
\begin{equation}
\delta \mathbf{z}_a = {\mathbf{R}^{n}_{b}\mathbf{a}}^{b} - \tilde{\mathbf{a}}^{n}
\end{equation}
where $\tilde{\mathbf{a}}^{n}$ is the acceleration estimated from GNSS measurements  and $\tilde{\mathbf{a}}^{b}$ is the corresponding acceleration expressed in the body frame.
The true acceleration in the navigation frame is given by
\begin{equation} \label{eq:acc_true}
\mathbf{a}^{n} = \mathbf{R}^{n}_{b} \, \mathbf{f}^{b}_{ib} + \mathbf{g}^{n}
\end{equation}
where 
\(\mathbf{f}^{b}_{ib}\) is the specific force vector expressed in the body frame, 
\(\mathbf{R}^{n}_{b}\) is the transformation matrix from body to navigation frame, and 
\(\mathbf{g}^{n}\) is the gravity vector expressed in the navigation frame.\\
\noindent Assuming small attitude errors \(\boldsymbol{\phi}^{n}\) and accelerometer bias \(\mathbf{b}_a\), we use~\cite{farrell2008aided}:
\begin{equation} \label{eq:num1}
\mathbf{R}^{n}_{b} = \hat{\mathbf{R}}^{n}_{b} \, (\mathbf{I} + [\boldsymbol{\phi}^{n} \times])   
\end{equation}
\begin{equation} \label{eq:num2}
\mathbf{f}^{b}_{ib} = \tilde{\mathbf{f}}^{b}_{ib} - \mathbf{b}_a.    
\end{equation}
Substituting \eqref{eq:num1}-\eqref{eq:num2} into the true acceleration vector \eqref{eq:acc_true} gives
\begin{equation}
\mathbf{a}^{n} = \hat{\mathbf{R}}^{n}_{b} \, (\mathbf{I} + [\boldsymbol{\phi}^{n} \times]) (\tilde{\mathbf{f}}^{b}_{ib} - \mathbf{b}_a) + \mathbf{g}^{n} 
\end{equation}
\noindent After some algebra we can define the acceleration measurement residual as:
\begin{equation}
\delta \mathbf{a}^{n} = \hat{\mathbf{R}}^{n}_{b} \, [\tilde{\mathbf{f}}^{b}_{ib} \times] \boldsymbol{\phi}^{n}
- \hat{\mathbf{R}}^{n}_{b} \mathbf{b}_a.
\end{equation}
\noindent The corresponding measurement matrix is:
\begin{equation}\label{H_a}
\mathbf{H}_{a} =
\begin{bmatrix}
\mathbf{0}_{3\times 3} &\mathbf{0}_{3\times 3} & \hat{\mathbf{R}}^{n}_{b} \, [\tilde{\mathbf{f}}^{b}_{ib} \times] & - \hat{\mathbf{R}}^{n}_{b} & \mathbf{0}_{3\times 3}
\end{bmatrix}.
\end{equation}
\noindent The measurement model for the position is:
\begin{equation}\label{H_p}
\mathbf{H}_{p} =
\begin{bmatrix}
\mathbf{I}_{ 3} &\mathbf{0}_{3\times 3} & \mathbf{0}_{3\times 3} & \mathbf{0}_{3\times 3} & \mathbf{0}_{3\times 3}
\end{bmatrix}.
\end{equation}
\noindent The overall measurement matrix comprising of the position and acceleration updates is constructed by combining \eqref{H_a} and \eqref{H_p}:
 \begin{equation} \label{eq:H}
 \mathbf{H} = \left[ \begin{array}{ccccc}
\mathbf{I}_{ 3} &\mathbf{0}_{3\times 3} & \mathbf{0}_{3\times 3} & \mathbf{0}_{3\times 3} & \mathbf{0}_{3\times 3} \\
\mathbf{0}_{3\times 3} &\mathbf{0}_{3\times 3} & \hat{\mathbf{R}}^{n}_{b} \, [\tilde{\mathbf{f}}^{b}_{ib} \times] & - \hat{\mathbf{R}}^{n}_{b} & \mathbf{0}_{3\times 3}
\end{array} \right]. 
\end{equation}
\section{Analysis and Results}\label{res}
\subsection{Datasets}
\begin{figure}[h]
    \centering
    \includegraphics[width=1.0\linewidth]{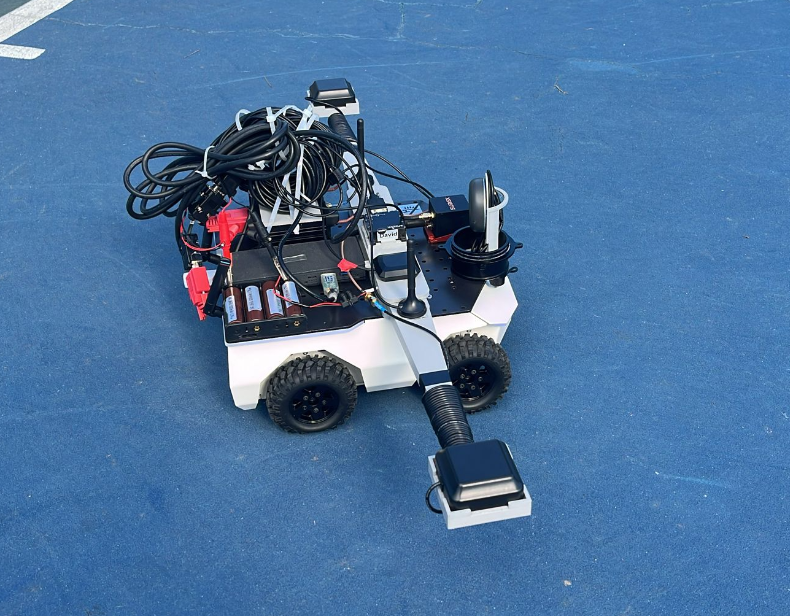}
 \caption{ Our recorded dataset setup using the ROSbot-Xl platform~\cite{husarion_rosbot_xl_manual} and Arazim's IMU EX-300 \cite{arazim_ex300}.}
    \label{platform_datasets}  
\end{figure}

\begin{figure}
    \centering
    \includegraphics[width=1.0\linewidth]{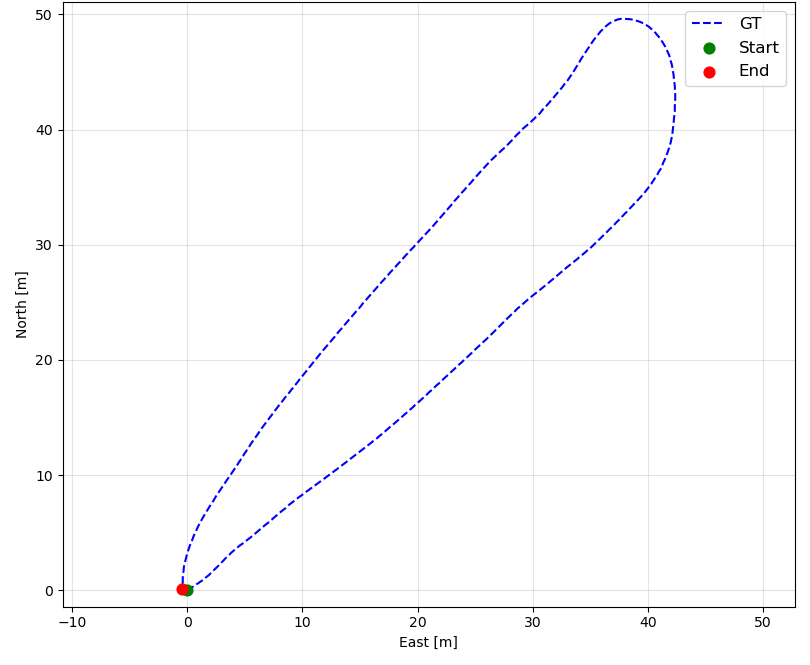}
    \caption{Trajectory number 1 of ROOAD dataset.}
    \label{traj_1_ofroad}
\end{figure}

\begin{figure}
    \centering
    \includegraphics[width=1.0\linewidth]{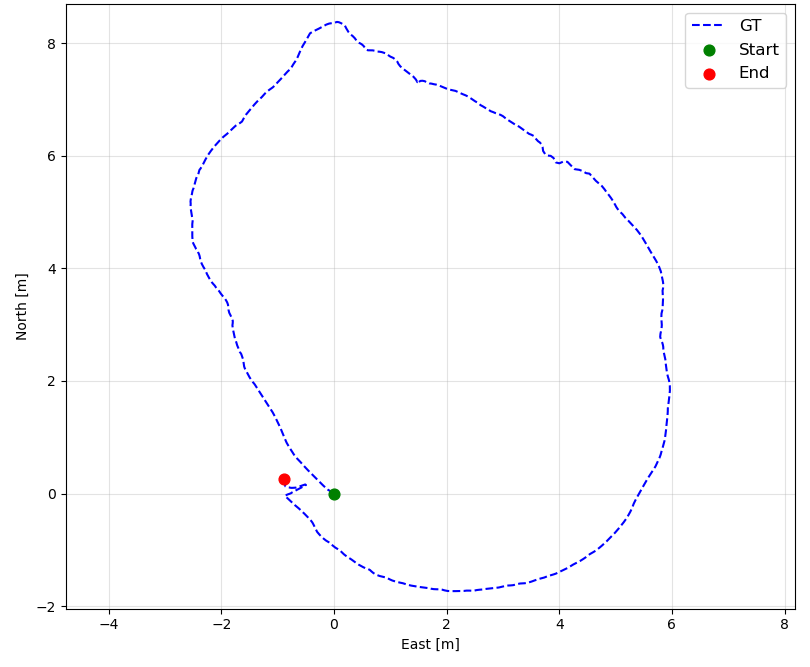}
    \caption{Trajectory number 8 of our dataset.}
    \label{traj_8_arazim}
\end{figure}

\noindent To evaluate our proposed approach, we utilized two real-world datasets collected from different mobile platforms, one of which is publicly available, while the other was recorded specifically for this study:
\begin{enumerate}
\item \textbf{ROOAD dataset:} This dataset~\cite{chustz2021rooad} was captured using the Waterhog-UGV platform~\cite{clearpath_warthog}, which is equipped with a VectorNav VN-300 INS~\cite{vectornav_vn300} and an Ardusimple dual-antenna GNSS receiver with RTK capability for precise position and heading estimation. The dataset spans approximately 45 minutes.

\item \textbf{Our dataset:} We recorded this dataset using a Rosbot-XL mobile robot platform~\cite{husarion_rosbot_xl_manual}, equipped with an Arazim Exiguo\textsuperscript{\textregistered} EX-300 IMU~\cite{arazim_ex300} and a dual-antenna GNSS system, allowing direct heading estimation. The recording lasted approximately 15 minutes.
\end{enumerate}
\noindent Altogether, the two datasets provide 60 minutes of real world recorded inertial and ground-truth (GT) measurements from two different mobile platforms, covering  distinct types of inertial sensors. Figure \ref{platform_datasets} present our plarform, sensors, and setup. Figures~\ref{traj_1_ofroad} and~\ref{traj_8_arazim} illustrate representative trajectories from each dataset.
\subsection{Performance Metric}
\noindent To evaluate the proposed approach we used a performance metric, as follows: 
\begin{enumerate}
    \item \textbf{Position root mean square error (PRMSE)}: The PRMSE of the 3D position compares the estimated position of the vehicle in the navigation frame with the GNSS-RTK GT position:
    \begin{equation} \label{eq:rmse}
    \text{PRMSE (m)} = \sqrt{\frac{1}{N} \sum_{k=1}^{N} ||\textbf{p}_k-\hat{\textbf{p}}^{}_k||^{2}}
    \end{equation}
    where $\hat{\textbf{p}}_k$ are the 3D estimated position vector at time k and $\textbf{p}_k$ is the GT position vector at time k.
\end{enumerate}
\subsection{Results}
\begin{figure}[h]
    \centering
    \includegraphics[width=1\linewidth]{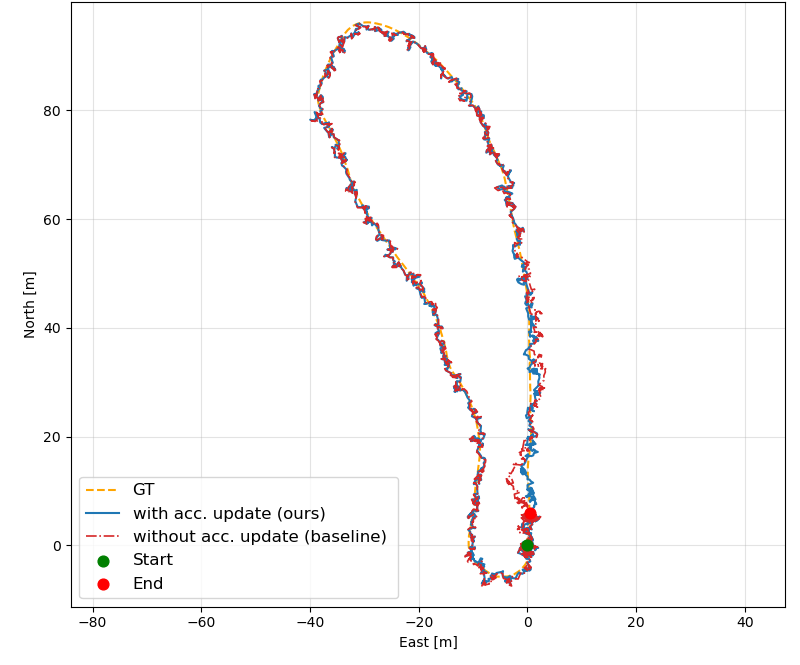}
    \caption{Trajectory number 4 with with acc update, without acc update, and the GT.}
        \label{error_pos_1}
\end{figure}

\noindent We compare our proposed approach, GNSS/INS with position and acceleration updates (with Acc, ours) to the baseline, GNSS/INS with only position updates (no Acc, baseline). 
\noindent The performance of our proposed method on the ROOAD dataset is summarized in Table \ref{prmse_rooad}. The baseline mean PRMSE of 5.00 m is reduced to 4.39 m by adding the acceleration updates, yielding a mean accuracy improvement of 11.40[\%]. Particularly noteworthy is Trajectory 3, which exhibited the highest error in the baseline (7.15 [m]) but saw a significant reduction to 5.97 [m]. Figure \ref{error_pos_1} shows the error position for each of the approaches over Trajectory 1.\\
\begin{table}[h]
	\centering
	\caption{PRMSE results for the ROOAD dataset.}
	\label{prmse_rooad}

	\resizebox{1.0\linewidth}{!}{
	\begin{tabular}{|c|c|c|c|}
		\hline
        \textbf{Trajectory} & \textbf{No Acc (baseline)} & \textbf{With Acc (ours)} & \textbf{Improvement {[}\%{]}} \\ 
        \hline
		\textbf{1}          & 4.84 & 4.28 & 11.60 \\ \hline
		\textbf{2}          & 3.46 & 3.21 & 7.20  \\ \hline
		\textbf{3}          & 7.15 & 5.97 & 16.40 \\ \hline
		\textbf{4}          & 4.54 & 4.05 & 10.80 \\ \hline
		\textbf{5}          & 3.75 & 3.45 & 7.80  \\ \hline
		\textbf{6}          & 6.28 & 5.36 & 14.60 \\ \hline
		\textbf{Average}       & \textbf{5.00} & \textbf{4.39} & \textbf{11.40} \\ \hline
	\end{tabular}
	}
\end{table}
\noindent Table \ref{prmse_arazim} presents PRMSE results for our dataset, evaluating the impact of acceleration data on trajectory estimation. The results demonstrate a consistent enhancement in accuracy when acceleration is integrated into the model as external update. Trajectory 9 showed the most significant improvement, with the error reducing from 4.43 [m] to 3.37 [m], representing a 23.9[\%] improvement.  The overall system achieved a mean PRMSE reduction of 20.74[\%], bringing the average error down to 3.44 m. 
\begin{table}[h]
\centering
\caption{PRMSE results for our dataset.}
\label{prmse_arazim}

\scriptsize
\resizebox{1.0\linewidth}{!}{
\begin{tabular}{|c|c|c|c|}
\hline
        \textbf{Trajectory} & \textbf{No Acc (baseline)} & \textbf{With Acc (ours)} & \textbf{Improvement {[}\%{]}} \\ 
        \hline
\textbf{1} & 4.27 & 3.37 & 21.00 \\ \hline
\textbf{2} & 4.26 & 3.40 & 21.1  \\ \hline
\textbf{3} & 4.36 & 3.34 & 21.30 \\ \hline
\textbf{4} & 4.51 & 3.45 & 23.60 \\ \hline
\textbf{5} & 4.39 & 3.52 & 19.80 \\ \hline
\textbf{6} & 4.14 & 3.51 & 15.20 \\ \hline
\textbf{7} & 4.33 & 3.52 & 18.60 \\ \hline
\textbf{8} & 4.35 & 3.51 & 19.40 \\ \hline
\textbf{9} & 4.43 & 3.37 & 23.9  \\ \hline
\textbf{Average} & \textbf{4.34} & \textbf{3.44} & \textbf{20.74} \\ \hline
\end{tabular}
}
\end{table}

\section{Conclusion}\label{conc}
\noindent This paper presented a method to improve INS/GNSS navigation 
filter accuracy by using the standard GNSS position updates with a GNSS-derived acceleration update. The acceleration estimate was obtained by applying a least-squares fitting over a sliding window of three past raw GNSS position measurements, requiring no additional sensors or hardware modifications. A linearized measurement model was derived, explicitly coupling the acceleration update to the orientation and inertial error states within the error-state EKF framework.
\noindent The proposed approach was evaluated on two real-world UGV 
datasets collected from different mobile platforms with different inertial sensors. Results consistently demonstrated improved positioning accuracy, with a mean PRMSE reduction of 11.40\% on the ROOAD dataset and 20.74\% on the our mobile robot dataset. The most significant improvements were observed in trajectories with higher baseline errors, suggesting that the acceleration update is particularly 
beneficial in challenging motion conditions.

\noindent In future work we aim to address the measurement correlation between the position update to the acceleration update, as the latter is constructed from past position history. To conclude, our approach offers a practical, hardware-free enhancement to standard INS/GNSS fusion, making it well-suited for real-time deployment on autonomous ground vehicles operating in diverse environments.
\section{Acknowledgment}
\noindent
The authors would like to thank Arazim Ltd. for their support and for providing the Arazim Exiguo EX-300 unit used in the data collection.

\bibliographystyle{ieeetr}
\bibliography{bio}

\end{document}